# CLASSIFICATION AND COMPARISON OF LICENSE PLATES LOCALIZATION ALGORITHMS

Mustapha Saidallah, Fatimazahra Taki, Abdelbaki El Belrhiti El Alaoui and Abdeslam El Fergougui

Computer Networks and Systems Laboratory, Faculty of Sciences, Moulay Ismail University, Meknes, Morocco

*ABSTRACT*

*The Intelligent Transportation Systems (ITS) are the subject of a world economic competition. They are the application of new information and communication technologies in the transport sector, to make the infrastructures more efficient, more reliable and more ecological. License Plates Recognition (LPR) is the key module of these systems, in which the License Plate Localization (LPL) is the most important stage, because it determines the speed and robustness of this module. Thus, during this step the algorithm must process the image and overcome several constraints as climatic and lighting conditions, sensors and angles variety, LPs' no-standardization, and the real time processing. This paper presents a classification and comparison of License Plates Localization (LPL) algorithms and describes the advantages, disadvantages and improvements made by each of them.*

## 1. INTRODUCTION

The mobility needs of people and products make the road traffic more and more cumbersome. ITS are designed to circumvent the challenges engendered by this problem. LPR is the centerpiece of this domain. It is based on three phases: the LPL, the characters segmentation of LP, and the Optical Characters Recognition (OCR).

An efficient LPL algorithm should provide accurate results in real time. Therefore, it must be able to manage the localization phase constraints: the multi-localization of all LP types, the noise due to climatic conditions, dirty LP or complex acquisition scenes, the lighting conditions that cause low contrast on the image or a multiple color appearance the non-standardization, the sensor/vehicle distance and the capture angle that significantly influence on colors, sizes and shapes of the LPs.

In our study, we classified the LPL algorithms into three families: the family of algorithms based on mathematical processing is presented in section 2, the family of algorithms based on the plate features is presented in section 3 and the family of algorithms based on deep learning in section 4. Comparative mapping performance of the studied algorithms according to this phase constraints is presented in section 5. In Section 6, a conclusion of our work will be provided.

## 2. FAMILY BASED ON MATHEMATICAL PROCESSING

The mathematical processing of images is a very active field, where the advanced theoretical are concretized to manipulate the image content. It consists in transforming an image to another to extract useful information for the application. It involves partial differential equations, geometric tools, and optimization concepts. Then for the validation phase the LP properties are exploited. This family can be divided into five algorithm classes based on: edges extraction, morphological operations, hybrid methods, transforms and Sliding Concentric Windows (SCW).

### 2.1. Class of Algorithms based on Edge Extraction

These algorithms apply the convolution of high-pass filter and the original image; they highlight the high frequencies and attenuate the rest. These are the techniques that transform the image into a set of edges, not necessarily closed, forming the significant borders of image. These methods consider the characters and the LP contours as a reference point for localization. They proceed essentially in four steps:

- The Application of filter to extract the edge points.
- The binarization of the Contours image by choosing a threshold.
- The Analysis of result to extract the candidates.
- The validation by geometrical, textural or density properties.

The authors of [1] use the Sobel filter to extract edges from the color image, and the histogram to select the binarization threshold. The logical AND of the binary and the HSV images, was exploited to extract the yellow candidates, and the other types are extracted from the Integral Image. The validation phase was based on the analysis of geometrical properties of Connected Component (CC). The regions with the high vertical gradient density are the candidates on [2]. The validation phase was based on aspect ratio, shape and tracking criterion.

In [3,4] the edge extraction is performed on the grayscale image to reduce the processing time and the texture was the criterion used to exactly isolate the LP.

The edge extraction in [5,6] is performed by the Canny filter, it is a noise reduction by a Gaussian filter before applying the Sobel filter. The contours density is the property that has confirmed the LP region.

Simplicity and speed are the qualities of these methods. But these qualities are defects in case of noisy images, complex acquisition scenes or if the LP borders do not present a great variation compared to the rest of image. Moreover, these algorithms use two thresholds, the edge extraction threshold which is difficult to be automatically selected under the various lighting conditions and the region contour density threshold for the validation phase that makes these algorithms very dependent to the sensor-vehicle distance.

### 2.2. Class of Algorithms based on Morphological Operations

This a set of nonlinear operators intended to probe the image with small geometries called Structuring Elements (SE) to highlight certain shapes in the image.

- The choice of the adequate structuring element.
- The application of morphological operations based on Dilation and Erosion.

- A binary representation of the pixels (white present the objects and black present the background).
- The CC Analysis based on the histogram of the results.

The authors of [7] describe a localization algorithm based on opening. It is a dilation of the erosed grayscale image. The validation phase was based on the horizontal and vertical projection of the result and the priori knowledge of the LP location. Therefore, the multilocalization is not allowed by this algorithm.

In [8] a combination of hat transform and morphological gradient was used to extract the candidates. The density, the location and the length/width ratio are used to confirm the LP region.

The strength of this class is its robustness to noise since it puts in advance the image shapes and denoises it at a time. Nevertheless, its treatments take an enormous time what makes them inadaptable to real-time systems. Moreover, the morphological operations don't give good results in case of poorly contrasted images where the sensitivity of these methods to the lighting conditions. Finally, the choice of adequate SE is an important parameter influenced and limited by a fixed sensor-vehicle distance.

## 2.3. Class of Hybrid Algorithms

These algorithms combine the edge detection and the morphological operations. They simplify the image by keeping just edges. Then they apply morphological operations to denoise the picture and complete the shapes, to define the CC. and based on geometrical and/or textural criteria the LPs are validated.

In [9] the magnitude of the vertical gradients is used to detect candidate regions. The opening is the morphological operation applied to reduce the noise and to remove the objects non-LP. Then the length/width ratio, the size and the orientation of candidates are the geometrical criteria used to validate the LP.

Treatment was reversed in [10] to solve the sensitivity of this class to lighting conditions. The algorithm starts with a Top-hat and Bottom-hat transforms that allowed to achieve a high contrasted grayscale image. Then the edges are extracted via the Sobel filter. The validation step consists in the CC analysis according to geometric criteria.

The authors of [11,12] propose a candidate's location by extracting the contours of the binary image using the Canny filter. They added to the opening a dilation operation to connect close objects. The region that checks the geometric criteria: rectangular shape and width/height ratio is a LP. Rotation of the plate is carried out in [10] for correcting the deviation.

In [13] the extraction of the image contours is performed by the second derivative via the Laplacian Of Gaussian (LOG) filter followed by a dilation to refine the contours and connect objects, and by classifying candidates to "text" and "no text", the LP is exactly located.

These algorithms showed a great improvement of the localization rate. Thus, they lend well to the no-standardization and the multi-localization constrain. However, the problems related to the edge extraction (Identification of thresholds under different lighting conditions) and to the morphological operations (Significant processing time) remain strongly posed.

## 2.4. Class of Algorithms based on Transforms

In this class we will study the sub-classes based on the Hough Transform (HT), the Wavelet Transform (WT) and Generalized Symmetry Transform (GST).

The Hough Transform is a technique of simple pattern recognition in an image by binding the pixels between them; its major disadvantage is the great processing time. To solve this problem, the authors of [14] proposed a pretreatment step which consists in binarizing the contour map of the original image, and on the result they applied the Hough Transform to extract the parallelogram objects. The validation phase consists in assessing the length/width ratio of candidates, then calculating the number of objects cut by two horizontal lines drawn at 1/3 and 2/3 of the candidate's height.

The HT does not give a good result when the image is noisy. The authors of [15] proposed to denoise the image via a Median filter to solve this problem. Then they applied a second filter to isolate the area where the plate is probably located to reduce the processing time.

The results of this subclass are accurate, and it can locate any type of plates at once, because it focuses on the shape of the plate and not on the type. However, these algorithms cannot be implemented in real-time systems or on images where the rectangular shape of the LP is not clear (Because of an obscuring object, continuity of intensity, or physical damage). Further, this type of algorithms cannot locate too deviated plates.

The Wavelet Transform decomposes the image to a set of cards, each presenting a characteristic. The authors of [16,17] started from the idea that the great difference in intensity between the LP and its characters is a key element to locate it. Therefore, they decompose the grayscale image into four frames via the WT, of which they used two. On the first, containing the horizontal details, they search the area where there is a maximum value of variation to locate a reference line. By analyzing the horizontal projection curve of the second image, containing the vertical details, they arrived to determine the sizes of the existing LPs on the image. Then by scanning the image they determined candidates. The validation phase was based on the geometric properties analysis.

These algorithms are independent of the sensor-vehicle distance and they are capable to detect several LP at the same time. However, they give false results when the contrast between the characters and the background is not quite clear. Thus, their sensitivity to the noises is high.

The GST in [18] evaluates evidence of local symmetry; it extracts a map of symmetrical contours of the image in all directions. The authors used in [19] the symmetry of the LP like an element-key to locate it. And to reduce the processing time the authors focused on the extraction of the four corners of the plate; consequently, they scanned the image only in the two diagonal directions (45 and -45). The multiplication of the two resultants and the connection of the extracted corners, give the LP candidates. Finally, a neural network system is used to validate the LP.

To adapt this subclass to real-time systems it is necessary to fix a limited number of scans. Thus, to achieve accurate results it is necessary to consider the lighting conditions and noise constraints. Moreover, the algorithm becomes weakening in the case of damaged LPs or if one of its corners is hidden.

## 2.5. Class of Algorithms based on Sliding Concentric Windows

This method is based on pixel intensities to extract the regions of interest. The LP localization in [20] passes by two stages: The first consists in scanning the image by two SCW using the standard deviation; if those values exceeds the predefined thresholds, the pixel belongs to a region of interest and it is posted at 1, if not the pixel is worth 0. The result is a binary image of candidates. The second phase uses geometrical criteria to validate the LP. The method is adapted to the multi-localization applications. However, since it is based on the pixel's intensities, a low-resolution or a poor lighting conditions are its blocking points. Thus, its simplicity made it unable to process noisy images or complex scenes.

## 3. FAMILY BASED ON LP FEATURES

These methods assume that a good prior knowledge of the LP properties, help to implement a robust model of its localization. They consider that the LP is an object that has defining properties in the image: specific texture, unique color composition, or both combinations. Identifying the LP properties is the first step of these algorithms. Then, each pixel is represented by a vector of these properties. Therefore, the classification methods classify and label pixels to indicate their affiliation or not to a region. Finally, the validation phase is based on geometrical criteria or histogram analysis. The localization rate of these methods is generally high, but they are specific to the treated LPs types.

### 3.1. Class of Algorithms based on LP Texture

The algorithms of this class are focused on the analysis of LP texture. They assume that the LP and its characters have a specific texture that distinguishes them from other image objects.
Based on the four following characteristics:

- In a small region contains more than one character,
- Characters and background are in sharp contrast,
- The size of the plate region is relatively fixed with a fixed length and width ratio,
- The tilt angle of license plate is in a certain range.

The algorithm in [21] constructs the attribute vectors to classify the pixels. The validation phase was based on the analysis of the result histogram.

The idea of [22] is that the LP is in a textural region of the image. Therefore, they decompose the image into blocks. The density of each block is calculated to classify it in "textural" or "not textural". The validation phase was based on geometrical criteria.

The authors of [23] implement four LPL algorithms based on the following LP features:

- The length/width ratio of the connected region,
- The background area,
- The character/plate area density,
- The number of edges in the plate center.

These features are exploited by deterministic, probabilistic and both combination methods, taking several thresholds for each one to validate the results.

This class uses the values showing the properties degree in a region. These values are directly affected by the noise, the sensor-vehicle distance, and the image resolution.

### 3.2. Class of Algorithms based on LP Colors

A unique colors combination which occurs almost exclusively in the LP area is the basic idea of these algorithms. In [24] the pixels classification is performed in the HSV (Hue, Saturation, Value) space to solve the problem of the wide RGB (Red, Green, Blue) value under different illumination. The validation phase was based on the evaluation of the length/width ratio and the projection curve analysis.

The authors of [25] have based their method on Fuzzy Logic. They decompose the color image into 4 cards: E denote the edge map computed using the color edge detector implemented by authors, H, S and V are the maps preserving the hue, saturation, and intensity components of the transformed image. E, H, S and V are the entries of the fuzzy system that classified the pixels. The result represents the degrees of the pixels belonging to a LP. The validation phase was delegated to the OCR. The richness of the test base confirmed the algorithm localization performance. However, the processing time made it inapplicable in real time systems.

The algorithms of this class reach high localization rates, but in an important processing time. Moreover, the sensitivity to the lighting conditions is their major disadvantage. Furthermore, the LPs have different colors and shapes because of the non-standardization which make these algorithms specifics to treated LP types.

### 3.3. Class of Algorithms based on Vector Quantization (VQ)

This concept is described in [26]. It is to define a set of vectors forming a dictionary. They include information on the image blocks. These blocks are obtained by selecting the high contrasted strips in a quad tree decomposition phase. These strips are validated by sorting them according to the expected LP size and the code words that describe the blocks content.

The authors of [26] assume that the LP corresponds generally to the first two bands of the sorted list. The test of this method in a real system has shown its high dependence to the sensor-vehicle distance (A considerable change in this factor requires the update of the entire system). Moreover, it can't process too noisy images or localize several plates at a time.

## 4. FAMILY BASED ON DEEP LEARNING

Deep learning is a new technique in the area of machine learning, which attempts to model high-level abstractions in data. There are various deep learning architectures such as deep convolutional neural network, deep belief network, and so on. Due to the deep network structures, they have been widely used in many applications with great success. Compared with other architectures, a deep convolutional neural network has fewer weights and less complexity. In addition, it allows one to directly use original images as inputs which enable a hierarchical learning of features.

The authors of [27] proposed a work-flow and methods for segmentation and annotation free Automated LPR (ALPR) with improved plate localization and automatic failure identification. Their proposed workflow first localizes the license plate region in the captured image using a two-stage approach where a set of candidate regions are first extracted using a weak SNoW classifier and then scrutinized by a strong convolutional neural network (CNN) classifier in the second stage. Images that fail a primary confidence test for plate localization are further classified to identify reasons for failure, such as license plate not present, LP too bright, LP too dark or no vehicle found. In the localized plate region, they perform segmentation and OCR jointly by using a probabilistic inference method based on hidden Markov models where the most likely code sequence is determined by applying the Viterbi algorithm. In order to reduce manual annotation required for training classifiers for OCR, they propose to use either artificially generated synthetic license plate images or character samples acquired by trained ALPR systems already operating in other sites. The performance gap due to differences between training and target domain distributions is minimized using an unsupervised domain adaptation. They evaluated the performance of their proposed methods on license plate images captured in several US jurisdictions under realistic conditions.

To avoid large accumulated error in typical three-step LPR methods and improve recognition performance in challenging complex environment, an end-to-end recognition method based on deep convolution neural network named LPR-Net is proposed in [28]. The input of LPR-Net is a gray or color image with any size and the output is the license plate number of input image. If no license plate in input images, LPR-net will output "no license". The proposed LPR-Net is a hybrid deep architecture that consists of a basic net, a multi-scale net, a regression net, and a classification net. Firstly, LPR-Net resizes the input image to $500 \times 500$ pixels. Then the resized image is input to basic net to get basic features. After that, the basic features are input to multiscale net to obtain multi-scale features. Finally, the multi-scale features are input to regression net and classification net to locate plate and identify characters.

The authors of [29] developed a deep learning application to detect and recognize Korean cars' license plates from images. It is an advanced application that targets to provide deep learning solution that can be applied in many areas including Intelligent Transportation System, Internet of Things and Smart City. They developed their method that is a combination of scene text recognition technique with Geometrical Image Transformation (GIT) to recognize number plates for combined neural networks and achieving 99.8% and 95.7% of detection and recognition accuracy respectively.

The authors of [30] proposed a novel machine learning approach to recognizing the license plate number. They used one of the most successful deep learning methods, convolutional neural network for extracting the visual features automatically. Suitable localization and segmentation techniques are employed before CNN model to enhance the accuracy of the proposed model. In addition to this, the Deep License Plate Number Recognition (D-PNR) model also takes care of proper identification from images those are hazy and is not suitable-inclined or noisy images. The efficiency of license plate recognition system improved using deep Learning framework to keep the trade-off between accuracy and time complexity. So, an efficient and real-time-based D-PNR model proposed for the license plate recrimination. It is able to give end results with 98.02% accuracy after setting suitable learning parameters.

The authors of [31] presented an embedded platform-based Italian license plate detection and recognition system using deep neural classifiers. In their work, trained parameters of a highly precise ALPR system are imported and used to replicate the same neural classifiers on an NVidia Shield K1 tablet. A CUDA-based framework is used to realize these neural networks. The flow of the trained architecture is simplified to perform the license plate recognition in realtime. Results show that the tasks of plate and character detection and localization can be performed in real-time on a mobile platform by simplifying the flow of the trained architecture. However, the accuracy of the simplified architecture would be decreased accordingly.

The authors of [32] used deep learning method for real-time traffic images. They first extract license plate candidates from each frame using edge information and geometrical properties, ensuring high recall using one class Support Vector Machines (SVM). The verified candidates are used for Number Plate (NP) recognition purpose along with a spatial transformer network (STN) for character recognition. Their system is evaluated on several traffic images with vehicles having different license plate formats in terms of tilt, distances, colors, illumination, character size, thickness etc. Also, the background was very challenging. Results demonstrate robustness to such variations and impressive performance in both the localization and recognition.

## 5. COMPARISON AND SYNTHESIS

The exactitude of a LPR system depends on the localization phase, i.e., to be efficient, its localization rate should be around 100%. The experimental tests of localization algorithms are not done on unified databases. Indeed, a rich database that allows to test the performance of the algorithms must contain many several plate images, taken in different lighting conditions, from various distances, angles, and acquisition scenes. These images must present all LP types. In addition, a localization algorithm should not simply isolate the LP, but it must also isolate it in real time.

To qualify the algorithms, we distinguish four levels of the "performance" parameter. This later binds the localization rate and the database richness as the Table 1 shows.

Indeed, if the database contains a significant number of images and various constraints related to this phase, its richness is very high and if the localization rate is around 100%, then the algorithm will be qualified "very powerful: AAA".

Table 1. Values of the performance parameter.

| Database richness | Localization rate | Performance |
|---|---|---|
| Very high | Very high | AAA |
| Very high | Medium | AA |
| Medium | High | AA |
| Medium | Medium | A |
| Low | High | A |
| High | Low | A |
| Low | Medium | B |
| Medium | Low | B |
| Low | Low | B |

In addition to the performance of the LPL algorithms, Table 2 shows their sensitivity to the localization phase constraints. This sensitivity is deduced from the analysis of algorithms and which is not necessarily present in the database. For example, if the database does not contain images of vehicles with constraints:

- LP type: the database contains only the car LP of a single line,
- No-standardization: the database contains only the plates respecting a given standard.

And if the analysis shows that the algorithm can support only one LP type of different standards, the answer in Table 2 will be "No" for the LP type and "Yes" for the Nostandardization.

From Table 2, we deduce on the one hand, the absence of an ideal LPL algorithm: very powerful and which solve the various constraints of this phase. Moreover, the most efficient algorithms are deep learning algorithms because they solve most constraints. On the other hand, the least overcome constraint by the different algorithms is sensitivity to lighting conditions. Thus, this constraint is circumvented just by nine algorithms; three of them are based on the LP features and four on deep learning.

Table 2. Performance of LPL algorithms and their sensitivity to localization constraints.

| Family | Class | | Algorithm | Performance | Real time | Multi-Localization | LP type | No-Standardization | Noise & Scene complexity | Lighting Conditions | Sensor/vehicle distance | Acquisition Angle |
|---|---|---|---|---|---|---|---|---|---|---|---|---|
| Mathematical processing | Edges extraction | | [1] | AAA | Yes | Yes | Yes | Yes | No | No | Yes | No |
| | | | [4] | AA | Yes | Yes | No | No | No | No | No | No |
| | | | [5] | A | Yes | No | Yes | No | No | No | No | No |
| | Morphological operations | | [7] | B | No | No | Yes | Yes | Yes | No | Yes | No |
| | | | [8] | B | No | No | Yes | Yes | Yes | No | Yes | No |
| | Hybrids | | [9] | A | Yes | No | No | No | Yes | No | No | No |
| | | | [11] | AAA | Yes | Yes | No | No | Yes | Yes | Yes | Yes |
| | | | [13] | B | Yes | Yes | Yes | No | No | No | Yes | Yes |
| | | | [10] | AAA | Yes | Yes | Yes | Yes | No | Yes | Yes | Yes |
| | Transforms | HT | [14] | AAA | Yes | Yes | No | Yes | No | No | No | No |
| | | WT | [16] | AA | Yes | No | Yes | Yes | No | No | No | No |
| | | GST | [18] | A | No | Yes | Yes | Yes | No | No | Yes | Yes |
| | SCW | | [20] | AAA | Yes | Yes | Yes | No | No | No | Yes | Yes |
| LP Features | Texture | | [21] | B | Yes | Yes | No | No | No | No | No | Yes |
| | | | [22] | B | Yes | Yes | No | No | No | Yes | Yes | Yes |
| | Color | | [24] | A | No | Yes | Yes | No | Yes | Yes | Yes | Yes |
| | | | [25] | AAA | No | Yes | Yes | Yes | Yes | Yes | Yes | Yes |
| | VQ | | [26] | B | No | No | Yes | Yes | No | No | No | Yes |
| Deep learning | | | [27] | AAA | Yes | Yes | Yes | No | Yes | Yes | Yes | No |
| | | | [28] | AA | Yes | Yes | No | Yes | Yes | Yes | Yes | Yes |
| | | | [29] | A | No | Yes | No | Yes | Yes | Yes | Yes | No |
| | | | [30] | AA | No | Yes | Yes | No | Yes | No | Yes | Yes |
| | | | [31] | AAA | Yes | Yes | No | Yes | Yes | Yes | Yes | No |
| | | | [32] | A | Yes | No | No | Yes | Yes | No | Yes | No |

## 6. CONCLUSIONS

We have presented the functioning principle of different LP Localization methods by analyzing the strengths and weaknesses of each of them. Our study has demonstrated the lack of an ideal localization algorithm and confirmed that a meaningful comparison of these methods performance requires their evaluation on the same test database, containing enough scenes with the various constraints of this phase. Finally, we fill a table showing main properties of the considered algorithms which could help a researcher to select the most appropriate one for his specific task and conditions.